%% file: main.tex
\documentclass[runningheads]{llncs}
\usepackage[T1]{fontenc}
\usepackage{graphicx,verbatim}
\usepackage{hyperref}
\usepackage{color}

\usepackage{subfiles}
\usepackage{booktabs}
\usepackage{ amssymb }
\usepackage{amsmath}
\graphicspath{ {./figures/} }
\usepackage{float}
\usepackage{multirow}
\usepackage{array}

\begin{document}
\title{Predictive Multimodal Modeling of Diagnoses and Treatments in EHR}
\author{Cindy Shih-Ting Huang \and
Clarence Boon Liang Ng \and
Marek Rei}

\authorrunning{C. Huang et al.}
%
\institute{Imperial College London, United Kingdom \\
\email{\{cindy.huang23, clarence.ng21\}@alumni.imperial.ac.uk}\\
\email{marek.rei@imperial.ac.uk}}

\maketitle              
\begin{abstract}
While the ICD code assignment problem has been widely studied, most works have focused on post-discharge document classification. Models for early forecasting of this information could be used for identifying health risks, suggesting effective treatments, or optimizing resource allocation. To address the challenge of predictive modeling using the limited information at the beginning of a patient stay, we propose a multimodal system to fuse clinical notes and tabular events captured in electronic health records. The model integrates pre-trained encoders, feature pooling, and cross-modal attention to learn optimal representations across modalities and balance their presence at every temporal point. Moreover, we present a weighted temporal loss that adjusts its contribution at each point in time. Experiments show that these strategies enhance the early prediction model, outperforming the current state-of-the-art systems.


\end{abstract}
\section{Introduction}

\subfile{sections/intro}

\section{Related Work}
\subfile{sections/related-work}

\section{Proposed Framework}
\subfile{sections/mm-rep}

\subfile{sections/framework}

\subfile{sections/training}

\section{Experiment Set-up and Results}
\subfile{sections/setup}

\subfile{sections/results}

\section{Conclusion}
\subfile{sections/conclusion}

\bibliography{custom, anthology}
\bibliographystyle{splncs04}

\end{document}

%% file: sections/intro.tex
Electronic health records (EHR) are comprehensive repositories of patient information, encompassing clinical notes, laboratory tests, diagnostic imaging, and other data sources that collectively document the medical trajectory of a given patient. Much of the research on EHR documents has focused on assigning accurate International Classification of Diseases (ICD) codes based on discharge summaries, which are written at the end of a hospital stay and contain a textual description of the relevant diagnoses and treatments \cite{mullenbach-etal-2018-explainable,Vu2020}. While recent work has shown that multimodal features \cite{Xu2019} and earlier clinical notes \cite{Ng2023} can provide additional useful context for this task, these studies have primarily aimed to automate the retrospective analysis of individual documents.

While ICD code classification during discharge has useful applications, the rich temporal structure of EHR has further potential. Systems for jointly modeling and predicting the overall health trajectory of a patient during hospitalization could potentially be used for identifying health risks, suggesting timely treatments, or optimizing healthcare workflow efficiency. The early assignment of diagnoses and treatments is a key factor in improving the effectiveness of patient care, yet very few works on EHR so far have explored prospective models that provide earlier prognostic estimates to allow for integration into clinical pipelines \cite{Ben-Israel2020}. Furthermore, no prior research has examined the impacts of incorporating multimodal information on the performance for this early-stage prediction task.

In this work, we investigate the use of multimodal learning to predict the diagnoses and treatments that patients will encounter. The system performs ICD code forecasts at various stages of the hospital stay, with the predictions continuously updated as more data becomes available. We design the \textbf{M}ultimodal \textbf{I}ntegrated \textbf{H}ierarchical \textbf{S}equence \textbf{T}ransformer (MIHST) architecture for augmenting the information in clinical notes with additional data sources, as these may reveal early indicators and complementary features which are not yet captured by textual reports. The model integrates pre-trained encoders, feature pooling, and cross-modal attention to learn optimal representations across modalities and balance their presence at every temporal point. 

Experiments show that this additional information is necessary for early prediction, as MIHST with textual and tabular data outperforms all existing models at any time cutoff prior to the final discharge summary. Cross-modal causal attention together with feature pooling is shown to be the best combination, as it allows the architecture to dynamically adjust to the shifting significance of each data source over time, eliminating the need for constant data availability or paired multimodal records. A novel loss function in the model also enhances early predictions by balancing the performance across multiple temporal points.
Code for the model and experiments are available at our repository.\footnote{\url{https://github.com/cindyellow/ehr-predictive-multimodal-modeling}}

%% file: sections/related-work.tex
The discharge summary has been a primary focus of research for automating ICD code assignments at the end of a stay. Initial models were based on convolutional neural networks (CNNs) \cite{mullenbach-etal-2018-explainable,liu-etal-2021-effective} and long short-term memory (LSTM) \cite{Vu2020,yuan-etal-2022-code}. Later, transformer approaches like the Pre-trained Language Model-ICD (PLM-ICD) \cite{Huang2022} and the Hierarchical Transformer for Document Sequences (HTDS) model \cite{Ng2023} surpassed their performance by dividing long documents into smaller sequences ("chunks") and retaining all token embeddings encoded to represent a document. Notably, HTDS also established the significance of including earlier clinical documents for improved ICD code classification, as these provide additional context for diagnoses and treatments.

Researchers have also investigated multimodal fusion to improve clinical task performance. Early fusion methods textualize other data types with associated source tags \cite{Niu2024} or inject token embeddings into the prompt via modal-specific encoders \cite{Belyaeva2024}. Recent work has also explored framing ICD code classification as a text-to-text task \cite{Boyle2023}, yet performance still lags behind state-of-the-art. These studies reveal limitations of early fusion, where textualization can obscure data properties and the relative priority of modalities. On the other hand, late fusion frameworks lack information flow between modalities, as seen by Xu et al. \cite{Xu2019}, who predict ICD-10 codes by averaging the outputs from separate models for notes and tabular events, relying on text availability when other modalities are missing.

These approaches were designed to output their prediction based on the discharge summary at the end of the hospital stay. In contrast, recent work has argued that for practical downstream applications, such code classification should instead be performed on earlier medical notes \cite{Cheng2023}.
The Label-Attentive Hierarchical Sequence Transformer (LAHST) \cite{Caralt2024} introduced temporal ICD code prediction using causal and label-wise attention for generating predictions at any time point, focusing only on the textual notes as input. 
Our proposed approach combines both textual and tabular information into a multimodal framework that allows for making real-time predictions throughout the hospital stay, improving performance during the crucial early stages with limited available evidence.

%% file: sections/mm-rep.tex
\subsection{Multimodal Representations} \label{sec:mm-rep}
\textbf{Tabular Feature Selection:} \label{sec:ft-selection} To assess the benefits of additional modalities, tabular events -- specifically laboratory measurements -- are examined in this work as they embody diverse information that can reveal valuable insights into disease progression, complementary to those mentioned in textual notes. These entries are represented as name-value pairs, where feature names denote event types and values are the corresponding measurements. Typically, measurement units and event entry time are also provided. We apply feature selection using the training and development sets to identify laboratory events most closely associated with ICD codes.

First, lab feature values undergo Yeo-Johnson transformation with standardization to ensure a uniform scale and Gaussian-like distribution. Missing values are imputed with the mean from the training set. We employ an iterative process of training a logistic regression model with an L1 penalty term for each ICD code. Models are trained on two variables per lab feature: the average measurement and the average difference between consecutive measurements in a stay. We start with features measured in more than 5,000 admissions. For each model, the 10 variables with the highest absolute coefficients are identified. We count how often each lab feature appears across all models and select those important for 20 or more ICD codes. ICD codes with micro-F1 scores (rescaled to 0-100) below 30 are retrained with an expanded variable set, achieved by lowering the admission threshold for filtering lab features to 2,000. The list of significant variables is updated for that code if its score improves. A second retraining phase targets codes with scores under 20, further reducing the threshold to 500 admissions. We do not further retrain to prevent overfitting to rare event types. The final list of 22 laboratory features used in the main model includes laboratory tests that are important for 10 or more ICD codes, as well as those significant for at least 5 labels among codes with scores less than 30.

\textbf{Tabular Representation:} To encode laboratory data, we employ the Tabular Prediction adapted BERT approach (TP-BERTa) \cite{Yan2024} pre-trained on classification tasks for a large tabular database. Measurements for events in the list of selected laboratory test features are normalized and discretized with a quantile bin value between 1 and 256 to align with the foundation model. Each bin value has been registered as \texttt{mask\_token\_id + bin\_value} in the model vocabulary. Both the feature name and bin value are encoded, yielding an embedding matrix:
\begin{equation}
    \mathbf{U}_i = [\mathbf{E}_{\text{CLS}}, \mathbf{E}_{\text{name}}^i, \mathbf{E}_{\text{value}}^i] \in \mathbb{R}^{(F + 2) \times D_\text{tabular}}
\end{equation}
where $F$ is the maximum number of tokens used to represent the feature name. $\mathbf{E}_{\text{value}}^i$ is scaled by the normalized lab measurement value.

After intra-feature attention, the embedding at the \texttt{[CLS]} position is used as the final representation of this feature, denoted $\mathbf{\hat{u}}_i \in \mathbb{R}^{1 \times D_\text{tabular}}$. The tensor of feature embeddings for all laboratory events is $\mathbf{E}_\text{tabular} = [\mathbf{\hat{u}}_1,\ldots, \mathbf{\hat{u}}_M] \in \mathbb{R}^{M \times D_\text{tabular}}$, where $M$ is the number of lab events for the admission and $D_\text{tabular}$ is the hidden dimension of TP-BERTa.

\textbf{Document Representation:} Documents are divided into chunks of $T$ tokens that can be encoded by a pre-trained language model (PLM). During training, a maximum of $N$ chunks are selected to limit resource usage. The resulting matrix $\mathbf{S} \in \mathbb{R}^{N \times T}$ contains all the chunk tokens that serve as input to the model. $\mathbf{S}$ is passed through the PLM to obtain token embeddings, where the tensor at the \texttt{[CLS]} position is selected to represent the document chunk, yielding $\mathbf{E}_\text{note} \in \mathbb{R}^{N \times D_\text{textual}}$. $D_\text{textual}$ is the hidden dimension of the PLM. We use \texttt{RoBERTa-base-PM-M3-Voc} as the PLM since it was pre-trained on abstracts and full-text content of biomedical works on PubMed and physician notes from MIMIC-III \cite{lewis-etal-2020-pretrained}.

%% file: sections/framework.tex
\begin{figure*}[t]
\centering
  \includegraphics[width=12cm]{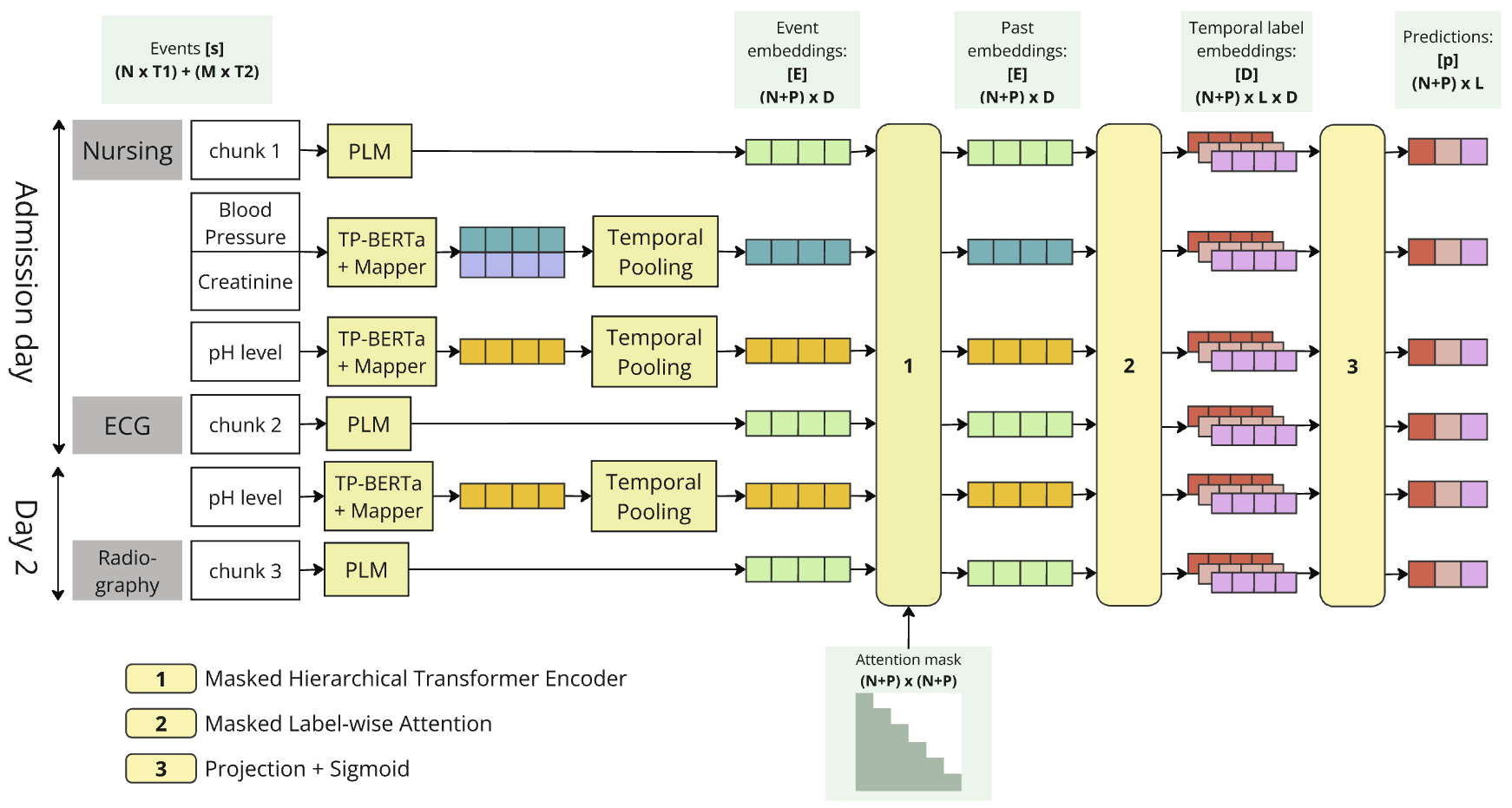}
  \caption{An overview of the model architecture. Document chunks are encoded by a pre-trained language model (PLM) while tabular measurements are encoded with TP-BERTa, followed by a modality mapper. The resulting tabular representations are pooled based on their timestamps. Updated encodings are passed through the masked hierarchical transformer and label-wise attention. Finally, a projection and sigmoid module outputs the label predictions.}
  \label{fig:model}
\end{figure*}

\subsection{Model Design}
We develop a model that integrates information from multiple modalities without requiring paired multimodal data for predictions during a patient stay. The model encodes chunks of medical documents and tabular event records, pooling tabular embeddings by timestamp. Textual and tabular representations are merged and sorted chronologically, then given as input to a hierarchical transformer. Next, a causally masked label-wise attention network extracts relevant information for each label up to that time point. Finally, label-specific embeddings are processed by a projection layer to generate temporal predictions for each ICD code. We refer to the model as the \textbf{M}ulti-modal \textbf{I}ntegrated \textbf{H}ierarchical \textbf{S}equence \textbf{T}ransformer (MIHST), illustrated in Figure \ref{fig:model}.

\textbf{Step 1: Clinical event encoding.}
As described in Section \ref{sec:mm-rep}, clinical events are encoded by either the PLM or the tabular foundation model, depending on the event type. Two embeddings are obtained: $\mathbf{E}_\text{note}$ for document chunks and $\mathbf{E}_\text{tabular}$ for laboratory measurements.

\textbf{Step 2: Modality mapper.} To align the tabular and textual vector spaces, $\mathbf{E}_\text{tabular}$ is updated with a trainable mapping network \cite{Ramos2024} consisting of a linear layer that transforms the tabular dimension $D_{\text{tabular}}$ to the textual dimension $D_{\text{textual}}$, followed by a LeakyReLU. 

\textbf{Step 3: Feature pooling.} Feature pooling enables the model to manage large volumes of tabular events without increasing computational resources or complexity. 

Embeddings with the same timestamp are pooled, condensing the tabular dimension into representations comparable to document chunk embeddings, which encapsulate textual information at a given time. For $p \in [1,\ldots,P]$, $P$ being the total number of unique temporal points for tabular events, feature pooling for the $p^{th}$ temporal position is performed on $\mathbf{w}(p)$, the set of tabular embeddings with time $p$. The entry time of the $m^{th}$ event is denoted $\textit{Time}(m)$.
\begin{equation}
\begin{aligned}
    \mathbf{E}^p_\text{pooled} &= \max_{m \in \mathbf{w}(p)}(\mathbf{E}^m_\text{tabular}) \\
    \mathbf{w}(p) = \{m &\in [1, \ldots, M] | \textit{Time}(m) = p \}
\end{aligned}  
\end{equation}

\textbf{Step 4: Causal Attention.} $\mathbf{E} = [\mathbf{E}_\text{note}, \mathbf{E}_\text{pooled}] \in \mathbb{R}^{(N+P) \times D_\text{textual}}$ is obtained by merging and sorting $\mathbf{E}_\text{note}$ and $\mathbf{E}_\text{pooled}$ by event timestamp. A hierarchical transformer with causal attention \cite{Choromanski2022} refines event embeddings with information from prior events. A masked attention block ensures each position accesses only past information. This generates an embedding matrix $\mathbf{H} \in \mathbb{R}^{(N+P) \times D_\text{textual}}$, where $H_i = \text{CausalAttn}(e_1, \ldots, e_i), i\in[1,\ldots,N+P]$.

\textbf{Step 5: Masked label-wise attention.} Label-wise attention network \cite{mullenbach-etal-2018-explainable} is utilized to prevent any predictions based on future events. The mask at temporal point $t$, denoted $a_t$, is constant in the label dimension and nullifies events beyond $t$. Multi-head attention \cite{Vaswani2017} with learnable label embeddings $\mathbf{Q} \in \mathbb{R}^{L \times D_\text{textual}}$ is applied. Linear projections of the key, query, and value embeddings $e_{k,i} = \mathbf{H}\mathbf{W}_i^K$, $e_{q,i} = \mathbf{Q}\mathbf{W}_i^Q$, $e_{v,i} = \mathbf{H}\mathbf{W}_i^V$ are used for each head. The output is $\mathbf{D}_t = \text{MultiHeadAttn} (\mathbf{Q}, \mathbf{H}, \mathbf{H}, {a}_t) \in \mathbb{R}^{1 \times L \times D_\text{textual}}$, which are label-specific embeddings for $L$ labels at each time point $t \in [1,\ldots,N+P]$.

Lastly, $\mathbf{D}_{t,\ell} \in \mathbb{R}^{D_\text{textual} \times 1}$ is passed through a projection layer followed by the sigmoid function: $p_{t,\ell} = \text{Sigmoid}(\mathbf{W}_\ell \cdot \mathbf{D}_{t,\ell})$. This represents the probability for the $l^{th}$ label at time $t$. Masking in the preceding modules guarantees that each output is computed using only embeddings of past events.

%% file: sections/training.tex
\textbf{Training:} The same training scheduler and hyperparameters as LAHST \cite{Caralt2024} are used, tuning only the temporal loss weights. We similarly apply the Extended Context Algorithm (ECA) for textual documents to accommodate indefinite document length by randomly sampling a maximum of $N_\text{max}$ text chunks during training. Tabular events do not face the same restraint during the encoding step, so we retain all of them to minimize information loss.

During inference, ECA is adapted for multimodality. $N_{\text{total}}$ textual note chunks are processed in batches of size $N_\text{max}$. Tabular events between the earliest and latest notes in the batch are also included, with $M_i$ denoting the number of tabular entries in batch $i$. The model encodes both inputs into batch embeddings, which are concatenated to form the embedding for the entire sequence of clinical events $h \in \mathbb{R}^{(N_{\text{total}}+M) \times D_\text{textual}}$, $M = \sum M_i$. This is passed to the masked multi-head label attention module for predictions based on the entire event sequence. If no lab records are present for a sample, the model proceeds under the unimodal setting using textual information.

Model training employs binary cross-entropy loss, computed per label $l$ among $L$ total labels by comparing predictions $p_l$ with ground truth $y_l$. To enhance early performance, we consider the loss across a set of temporal points $C$, where labels are compared against predictions based on events up to each $t \in C$. We further propose a weighted temporal loss to adjust the contribution of each time point to gradient propagation: $\mathcal{L_\text{w}} = - \sum_{t \in C} w_t [\frac{1}{L}\sum_{\ell=1}^L (y_\ell \cdot logp_{t,\ell}) + ((1-y_\ell) \cdot log(1-p_{t,\ell})) ]$, where the weights $w_t$ sum to 1. $C$ in our setup includes 5 temporal points: 2, 5, 13 days after admission, the time point right before the discharge summary, and the entry time of the summary -- these temporal positions are also used in evaluation. Experiments showed the best results when the last temporal point is given the highest weight (0.6) and others assigned 0.1.

%% file: sections/setup.tex
\textbf{Dataset:} For this study, we use the MIMIC-III \cite{Johnson2016} dataset, which contains de-identified multimodal health records from patients admitted to critical care units (ICU) between 2001 and 2012 at the Beth Israel Deaconess Medical Center in Boston, Massachusetts. We align data preprocessing, train/development/test splits, and label space with previous studies \cite{mullenbach-etal-2018-explainable,Caralt2024} for comparability, using the top 50 most frequent codes for modeling and evaluation. Note that patients are not excluded if they lack measurements for those laboratory tests. 

\textbf{Evaluation Framework:} In this task, we define temporal cutoffs at 2 days, 5 days and 13 days for standardized comparison with LAHST \cite{Caralt2024}. For instance, in the 5-day setting, the model predicts ICD codes based on textual and laboratory events occurring within the first 5 days of admission. We also report the model performance using all events up to (but excluding) the discharge summary to test the model without it, and with all events including the summary. Metrics follow standard conventions in the ICD coding task \cite{mullenbach-etal-2018-explainable}.

\begin{table*}[t!]
\caption{Evaluation on the test set for early ICD code prediction using all data until each of the specified temporal cutoffs. TrLDC performance is from the original paper \cite{Dai2022}. PubMedBERT-Hier (PMB-H; \cite{Ji2021}), HTDS \cite{Ng2023}, and LAHST results are from \cite{Caralt2024}. HTDS* is a variation of HTDS with similar computation requirements as LAHST and MIHST. Results for all models except TrLDC are averaged across 3 runs with random seeds. Standard deviations for MIHST results are < 0.3.}
\label{tab:test-set}
\centering
\begin{tabular}{l|>{\centering\arraybackslash}p{0.065\textwidth}>{\centering\arraybackslash}p{0.065\textwidth}>{\centering\arraybackslash}p{0.065\textwidth}|>{\centering\arraybackslash}p{0.065\textwidth}>{\centering\arraybackslash}p{0.065\textwidth}>{\centering\arraybackslash}p{0.065\textwidth}|>{\centering\arraybackslash}p{0.065\textwidth}>{\centering\arraybackslash}p{0.065\textwidth}>{\centering\arraybackslash}p{0.065\textwidth}|>{\centering\arraybackslash}p{0.065\textwidth}>{\centering\arraybackslash}p{0.065\textwidth}>{\centering\arraybackslash}p{0.065\textwidth}}
\toprule
 & \multicolumn{3}{c|}{\textbf{Last day}} & \multicolumn{3}{c|}{\textbf{0-13 days}} & \multicolumn{3}{c|}{\textbf{0-5 days}} & \multicolumn{3}{c}{\textbf{0-2 days}}\\ \midrule
\textbf{Model} & F1 & AUC & P@5 & F1 & AUC & P@5 & F1 & AUC & P@5 & F1 & AUC & P@5 \\ \midrule
TrLDC  & 70.1 & 93.7 & 65.9 & - & - & - & - & - & - & - & - & -\\ 
PMB-H  & 67.2 & 91.5 & 63.0 & 30.7 & 68.0 & 30.2 & 31.3 & 68.4 & 31.0 & 31.7 & 68.7 & 31.5\\ 
HTDS  & \textbf{73.3} & \textbf{95.2} & \textbf{68.1} & 49.7 & 82.1 & 47.6 & 47.5 & 80.6 & 45.9 & 44.5 & 78.7 & 43.6\\ 
HTDS*  & 70.7 & 93.8 & 66.2 & 48.6 & 82.0 & 47.0 & 46.7 & 80.7 & 45.5 & 43.6 & 78.7 & 43.3\\ 
LAHST  & 70.4 & 94.7 & 67.6 & 52.9 & 87.0 & 52.8 & 50.3 & 85.3 & 50.5 & 46.1 & 82.9 & 46.9\\ 
MIHST & 66.0 & 93.5 & 65.0 & \textbf{53.1} & \textbf{88.1} & \textbf{54.2} & \textbf{51.6} & \textbf{86.9} & \textbf{52.5} & \textbf{48.2} & \textbf{85.0} & \textbf{49.9} \\
\bottomrule
\end{tabular}
\end{table*}

%% file: sections/results.tex
We compare MIHST to existing baselines for real-time prediction. TrLDC \cite{Dai2022}, PMB-H \cite{Ji2021}, and HTDS \cite{Ng2023} are the best-performing models for the post-discharge task, while LAHST \cite{Caralt2024} serves as the state-of-the-art for early predictions during hospitalization. As shown in Table \ref{tab:test-set}, MIHST consistently outperforms LAHST in early prediction settings, achieving higher Micro-F1, Micro-AUC, and Precision@5 scores. By integrating multimodal representations, MIHST leverages both textual and non-textual information to make more accurate early predictions -- this is especially important in the early stages of the hospital stay, as each individual modality contains very limited information. 

For post-discharge predictions, other approaches outperform MIHST, likely due to the trade-off from optimizing across multiple time points and modalities. This indicates that optimal model choice depends on the required application: MIHST excels in all early-stage prediction settings, while unimodal models learning from the discharge summary may be more effective for post-discharge assignments.

\textbf{Ablation Experiments:} Table \ref{tab:ablation} presents key ablation results. Removing pooling lowers performance, indicating its role in preserving the strongest signals and preventing overfitting. We also examine the impact of weighted temporal loss across three settings: "Equal" assigns uniform weights to all time points; "First" assigns the largest weight (0.6) to the first cutoff and 0.1 to the rest; and "None" removes temporal loss, optimizing only for last-day predictions. The "Equal" setting maintains early performance but decreases the last-day score, likely due to reduced emphasis on the discharge summary, which only appears in the last temporal point and is the most relevant evidence for ICD codes. The low scores in the "First" setup further highlight the value of the summary in complementing other clinical records. Nevertheless, the "None" setting shows that distributing weights across all time points is crucial to enhance early prediction.

\begin{table}[t!]
\caption{Micro-F1 scores computed on the development set, comparing with feature pooling ablated and different weight schemes. Values are averaged across 3 runs, with the standard deviation shown in the subscript.}
\label{tab:ablation}
\centering
\small
\begin{tabular}{l|>{\centering\arraybackslash}p{0.13\textwidth}>{\centering\arraybackslash}p{0.15\textwidth}>{\centering\arraybackslash}p{0.13\textwidth}>{\centering\arraybackslash}p{0.13\textwidth}>{\centering\arraybackslash}p{0.13\textwidth}}
\toprule
& Model & Abl. Pooling & Equal & First & None\\ \hline
2 day & \textbf{49.6} \textsubscript{$\pm$0.1} & 48.8 \textsubscript{$\pm$0.5} & 49.4 \textsubscript{$\pm$0.3} & 48.0 \textsubscript{$\pm$0.3} & 47.8 \textsubscript{$\pm$0.5}\\
5 day & \textbf{54.0} \textsubscript{$\pm$0.3} & 52.8 \textsubscript{$\pm$0.6} & 52.8 \textsubscript{$\pm$0.2} & 51.0 \textsubscript{$\pm$0.4} & 52.2 \textsubscript{$\pm$0.5}\\
13 day & \textbf{56.1} \textsubscript{$\pm$0.3} & 54.7 \textsubscript{$\pm$0.5} & 54.7 \textsubscript{$\pm$0.2} & 52.1 \textsubscript{$\pm$0.3} & 54.7 \textsubscript{$\pm$0.6} \\
Excl. DS & \textbf{56.4} \textsubscript{$\pm$0.4} & 55.0 \textsubscript{$\pm$0.5} & 55.0 \textsubscript{$\pm$0.3} & 52.2 \textsubscript{$\pm$0.4} & 55.0 \textsubscript{$\pm$0.5} \\
Last day & 68.6 \textsubscript{$\pm$0.3} & 67.4 \textsubscript{$\pm$0.5} & 64.9 \textsubscript{$\pm$0.2} & 61.1 \textsubscript{$\pm$0.3} & \textbf{69.5} \textsubscript{$\pm$0.3}\\ 
\bottomrule
\end{tabular}
\end{table}

%% file: sections/conclusion.tex
This study leverages multimodal data to predict ICD codes at various points during hospitalization, with an emphasis on early prediction of diagnoses and treatments for 
a given patient. Strengthening prediction quality at the beginning of a hospital stay has the potential to aid clinicians in improving patient outcomes and planning resources. MIHST utilizes pre-trained foundation models for meaningful textual and tabular encodings, which then interact in a causal attention module that updates each representation based on previous information. A weighted temporal loss contributes to an optimal balance between predictions at temporal points.

Experiments demonstrated that multimodality benefits predictions when unimodal data offers weaker evidence, notably soon after admission. The novel weighted temporal loss aligns optimizations across temporal positions, while feature pooling moderates modality presence to emphasize the most informative features. This yields a system that surpasses the state-of-the-art for early predictions.

MIHST is agnostic to the pre-trained encoder choice and easily extends to new modalities via modality-specific encoders. Its cross-modal interaction design adapts to varying data availability and alignment. As more powerful PLMs emerge, MIHST can leverage a wider range of data sources to improve diagnoses and treatment decisions. The results highlight the potential for mining rich multimodal EHR data to advance prospective applications in clinical practices.